\documentclass[10pt,twocolumn,letterpaper]{article}
\usepackage{iccv}
\usepackage{times}
\usepackage{epsfig}
\usepackage{graphicx}
\usepackage{epstopdf}
\usepackage{amsmath}
\usepackage{amssymb}
\usepackage[colorlinks,linkcolor=black,anchorcolor=black,citecolor=green]{hyperref}

\iccvfinalcopy % *** Uncomment this line for the final submission

\def\iccvPaperID{1345} % *** Enter the ICCV Paper ID here

\ificcvfinal\fi
\begin{document}

%%%%%%%%% TITLE
\title{Detail-revealing Deep Video Super-resolution}
\author{Xin Tao\footnotemark[1] \quad Hongyun Gao\footnotemark[1] \quad Renjie Liao\footnotemark[2] \quad Jue Wang\footnotemark[3] \quad Jiaya Jia\footnotemark[1]\\
\footnotemark[1]~~The Chinese University of Hong Kong\\
\footnotemark[2]~~University of Toronto\quad\footnotemark[3]~~Megvii Inc.\\
%{\tt\small \{xyshen,hygao,xtao,zhouc,leojia\}@cse.cuhk.edu.hk}
}

\maketitle
%\thispagestyle{empty}

%%%%%%%%% ABSTRACT
\begin{abstract}
Previous CNN-based video super-resolution approaches need to align multiple frames to the
reference. In this paper, we show that proper frame alignment and motion compensation is
crucial for achieving high quality results. We accordingly propose a ``sub-pixel motion
compensation'' (SPMC) layer in a CNN framework. Analysis and experiments show the
suitability of this layer in video SR. The final end-to-end, scalable CNN framework
effectively incorporates the SPMC layer and fuses multiple frames to reveal image details.
Our implementation can generate visually and quantitatively high-quality results, superior
to current state-of-the-arts, without the need of parameter tuning. 
\end{abstract}

\let\thefootnote\relax\footnotetext{Code will be available upon acceptance: \href{https://github.com/jiangsutx/SPMC_VideoSR}{link}}
%%%%%%%%% BODY TEXT
\vspace{-0.1in}
\section{Introduction}
As one of the fundamental problems in image processing and computer vision, video or
multi-frame super-resolution (SR) aims at recovering high-resolution (HR) images from a
sequence of low-resolution (LR) ones. In contrast to single-image SR where details have
to be generated based on only external examples, an ideal video SR system should be able to
correctly extract and fuse image details in multiple frames. To achieve this goal, two
important sub-problems are to be answered: (1) how to align multiple frames to construct
accurate correspondence; and (2) how to effectively fuse image details for high-quality outputs.

%It is expected to recover details by maximally utilizinginformation from neighboring input frames. Therefore, in previous work, two domain-specific issues need to be specially addressed, especially for those deep learning based methods.

%-------------------------------------------------------------------------
\vspace{-0.1in}
\paragraph{Motion Compensation}
While large motion between consecutive frames increases the difficulty to locate
corresponding image regions, subtle sub-pixel motion contrarily benefits restoration of
details. Most previous methods compensate inter-frame motion by estimating optical flow
\cite{caballero2016real,farsiu2004fast,liao2015video,liu2011bayesian,ma2015handling}
or applying block-matching \cite{takeda2009super}.
After motion is estimated, traditional methods
\cite{farsiu2004fast,liu2011bayesian,ma2015handling} reconstruct the HR output
based on various imaging models and image priors, typically under an iterative
estimation framework. Most of these methods involve rather intensive case-by-case
parameter-tuning and costly computation.

Recent deep-learning-based video SR methods \cite{caballero2016real,kappeler2016video}
compensate inter-frame motion by aligning all other frames to the reference one, using
backward warping. We show that such a seemingly reasonable technical choice is actually
not optimal for video SR, and improving motion compensation can directly lead to higher
quality SR results. In this paper, we achieve this by proposing a sub-pixel motion
compensation (SPMC) strategy, which is validated by both theoretical analysis  and
extensive experiments.

%-------------------------------------------------------------------------
\vspace{-0.1in}
\paragraph{Detail Fusion}
Besides motion compensation, proper image detail fusion from multiple frames is the key
to the success of video SR. We propose a new CNN framework that incorporates the SPMC
layer, and effectively fuses image information from aligned frames. Although previous
CNN-based video SR systems can produce sharp-edge images, it is not entirely clear
whether the image details are those inherent in input frames, or learned from external data.
In many practical applications such as face or text recognition, only true HR details are
useful. In this paper we provide insightful ablation study to verify this point.

\vspace{-0.1in}
\paragraph{Scalability} A traditionally-overlooked but practically-meaningful
property of SR systems is the scalability. In many previous learning-based SR systems,
the network structure is closely coupled with SR parameters, making them less flexible
when new SR parameters need to be applied. For example, ESPCN \cite{shi2016real} output
channel number is determined by the scale factor. VSRnet \cite{kappeler2016video} and
VESPCN \cite{caballero2016real} can only take a fixed number of temporal frames as input,
once trained.

In contrast, our system is fully scalable. First, it can take arbitrary-size input images.
Second, the new SPMC layer does not contain trainable parameters and can be applied for
arbitrary scaling factors during testing. Finally, the ConvLSTM-based
\cite{xingjian2015convolutional} network structure makes it possible to accept an
arbitrary number of frames for SR in testing phase.

\subsection{Related Work}
\paragraph{Deep Super-resolution}
With the seminal work of SRCNN \cite{dong2014learning}, a majority of recent SR methods
employ deep neural networks
\cite{dong2016accelerating,Johnson2016Perceptual,kim2016VDSR,kim2016DRCN,ledig2016photo,shi2016real}.
Most of them resize input frames before sending them to the network
\cite{dong2016accelerating,Johnson2016Perceptual,kim2016VDSR,kim2016DRCN}, and use very
deep \cite{kim2016VDSR}, recursive \cite{kim2016DRCN} or other networks to predict HR
results. Shi \etal \cite{shi2016real} proposed a subpixel network, which directly takes
low-resolution images as input, and produces a high-res one with subpixel location. Ledig
\etal \cite{ledig2016photo} used a trainable deconvolution layer instead.

For deep video SR, Liao \etal \cite{liao2015video} adopted a separate step to construct
high-resolution SR-drafts, which are obtained under different flow parameters. Kappeler
\etal \cite{kappeler2016video} estimated optical flow and selected corresponding patches
across frames to train a CNN. In both methods, motion estimation is separated from
training. Recently, Caballero \etal \cite{caballero2016real} proposed the first
end-to-end video SR framework, which incorporates motion compensation as a submodule.

\vspace{-0.1in}
\paragraph{Motion Estimation}
Deep neural networks were also used to solve motion estimation problems. Zbontar and
LeCun \cite{zbontar2016stereo} and Luo \etal \cite{luo2016efficient} used CNNs to learn a
patch distance measure for stereo matching. Fischer \etal \cite{fischer2015flownet} and
Mayer \etal \cite{mayer2015large} proposed end-to-end networks to predict optical flow
and stereo disparity.

Progress was made in spatial transformer networks \cite{jaderberg2015spatial} where a
differentiable layer warps images according to predicted affine transformation
parameters. Based on it, WarpNet \cite{kanazawa2016warpnet} used a similar scheme to
extract sparse correspondence. Yu \etal \cite{yu2016back} warped output based on
predicted optical flow as a photometric loss for unsupervised optical flow learning.
Different from these strategies, we introduce a {\it Sub-pixel Motion Compensation}
(SPMC) layer, which is suitable for the video SR task.

%-------------------------------------------------------------------------
\section{Sub-pixel Motion Compensation (SPMC)}
\label{sec:analysis}

\begin{figure}[ht]
\centering
    \includegraphics[width=.92\linewidth]{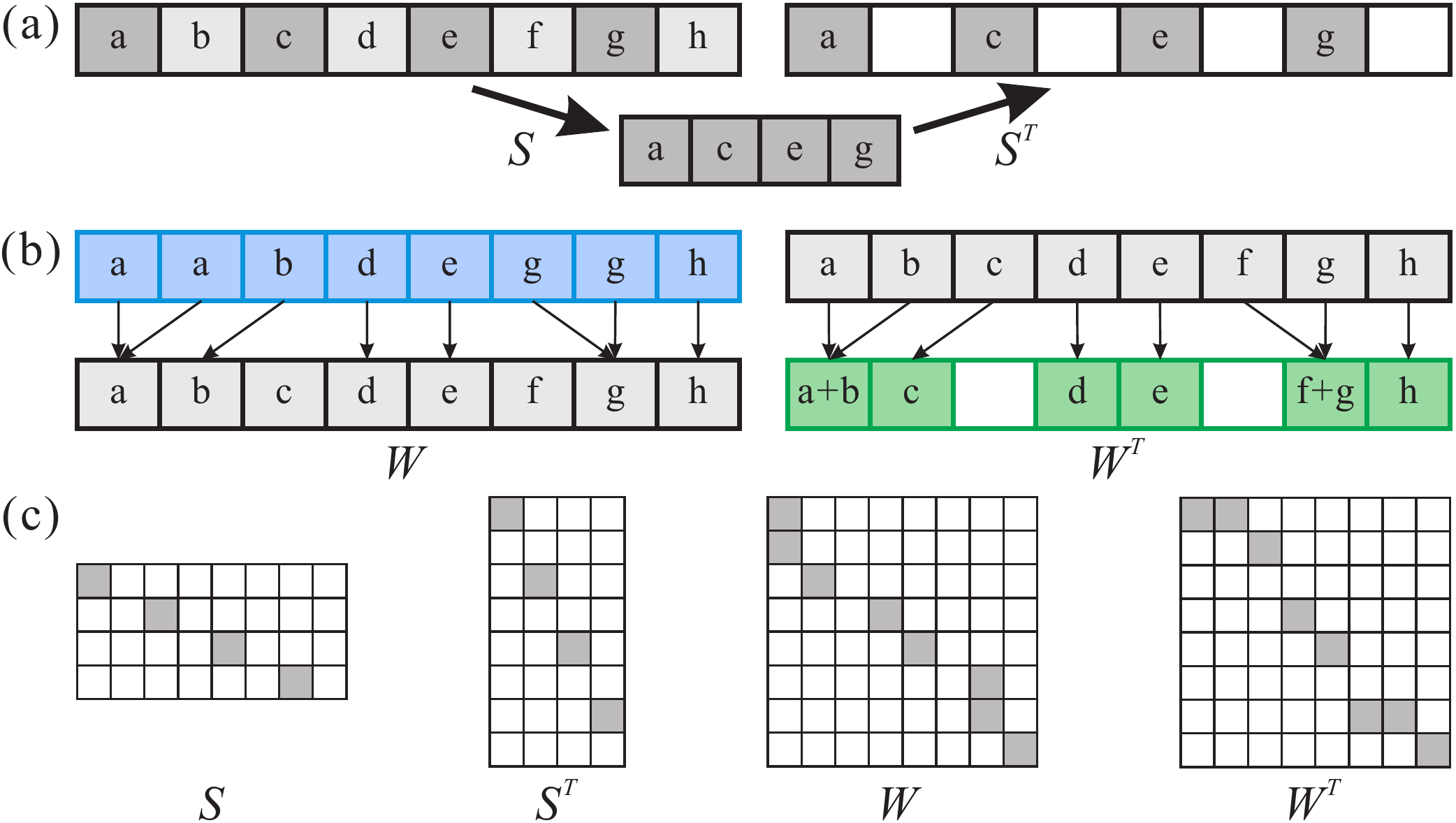}
\caption{\textbf{Visualization of operators in image formation}. (a) Decimation operator
$S$ ($2\times$) reduces the input 1D signal to its half-size. The transpose $S^T$
corresponds to zero-upsampling. (b) With arrows indicating motion, warping operator $W$
produces the {\bf blue} signal from the {\bf gray} one through backward warping. $W^T$
produces the {\bf green} signal through forward warping. (c) Illustration of matrices
$S$, $S^T$, $W$ and $W^T$. Grayed and white blocks indicate values 1 and 0 respectively.
}\label{fig:operators}
\end{figure}

We first introduce our notations for video SR. It takes a sequence of $N_F=(2T+1)$ LR
images as input ($T$ is the size of temporal span in terms of number of frames), where
$\Omega_L=\{I^L_{-T}, \cdots, I^L_{0}, \cdots, I^L_{T}\}$. The output HR image $I^H_0$
corresponds to center reference frame $I^L_{0}$.

\vspace{-0.1in}
\paragraph{LR Imaging Model}
The classical imaging model for LR images
\cite{farsiu2004fast,liao2015video,liu2011bayesian,ma2015handling} is expressed as
\begin{align}\label{eq:image_model}
    I^L_i = SKW_{0\rightarrow i} I^H_0+n_i,
\end{align}
where $W_{0\rightarrow i}$ is the warping operator to warp from the $0$th to $i$th frame.
$K$ and $S$ are downsampling blur and decimation operators, respectively. $n_i$ is the
additive noise to frame $i$. For simplicity's sake, we neglect operator $K$ in the
following analysis, since it can be absorbed by $S$.

\vspace{-0.1in}
\paragraph{Flow Direction and Transposed Operators}
Operator $W_{0\rightarrow i}$ indicates the warping process. To compute it, one needs to
first calculate the motion field $F_{i\rightarrow 0}$ (from the $i$th to $0$th frame),
and then perform backward warping to produce the warped image. However, current deep
video SR methods usually align other frames back to $I^L_0$, which actually makes use of
flow $F_{0\rightarrow i}$.

More specifically, directly minimizing the $L_2$-norm reconstruction error $\sum_{i}\Vert
SW_{0\rightarrow i} I^H_0-I^L_i \Vert^2$ results in
\begin{align}\label{eq:shift_and_add}
    I^H_0&=(\sum_i W_{0\rightarrow i}^T S^T S W_{0\rightarrow i})^{-1}(\sum_i W_{0\rightarrow i}^T S^T
    I_i^L).
\end{align}
With certain assumptions \cite{elad2001fast,farsiu2004fast}, $W_{0\rightarrow i}^T S^T S
W_{0\rightarrow i}$ becomes a diagonal matrix. The solution to
Eq.~(\ref{eq:shift_and_add}) reduces to a feed-forward generation process of
\begin{align}\label{eq:shift_and_add2}
    I^H_0&=\frac{\sum_i W_{0\rightarrow i}^T S^T I_i^L}{\sum_i W_{0\rightarrow i}^T S^T \mathbf{1}},
\end{align}
where $\mathbf{1}$ is an all-one vector with the same size as $I_i^L$. The operators that
are actually applied to $I_i^L$ are $S^T$ and $W_{0\rightarrow i}^T$. $S^T$ is the
transposed decimation corresponding to zero-upsampling. $W_{0\rightarrow i}^T$ is the
transposed forward warping using flow $F_{i\rightarrow 0}$. A 1D signal example for these
operators is shown in Fig.~\ref{fig:operators}. We will further analyze the difference of
forward and backward warping after explaining our system.

%\begin{figure}[t]
%\centering
%    \includegraphics[width=1.0\linewidth]{fig/diff_warp.pdf}
%\caption{\textbf{Baseline results by different warping}.
%    (a) Input (bicubic $\times4$).
%    (b) Baseline method: using $W$. (c) Baseline method: using $W^T$.
%    \xtao{caption change. figure enlarge, not that obvious. description ambiguous.}}\label{fig:diff_warp}
%\end{figure}

%The underlying principle can also be understood very intuitively. The process in
%Eq.~(\ref{eq:shift_and_add2}) is just moving each pixel of $I_i^L$ to an enlarged image
%grid. Since flow can have sub-pixel accuracy (floating numbers), pixels from $I^L_i$ have
%chance to fill in the empty intervals of HR grid of $I^H_0$. A visual example is shown in
%Fig.~\ref{fig:module_spmc}. On the other hand, previous motion compensation loses such
%sub-pixel information and thus cannot recover enough details.

%-------------------------------------------------------------------------
\section{Our Method}

\begin{figure*}[ht]
\begin{center}
\includegraphics[width=1.0\linewidth]{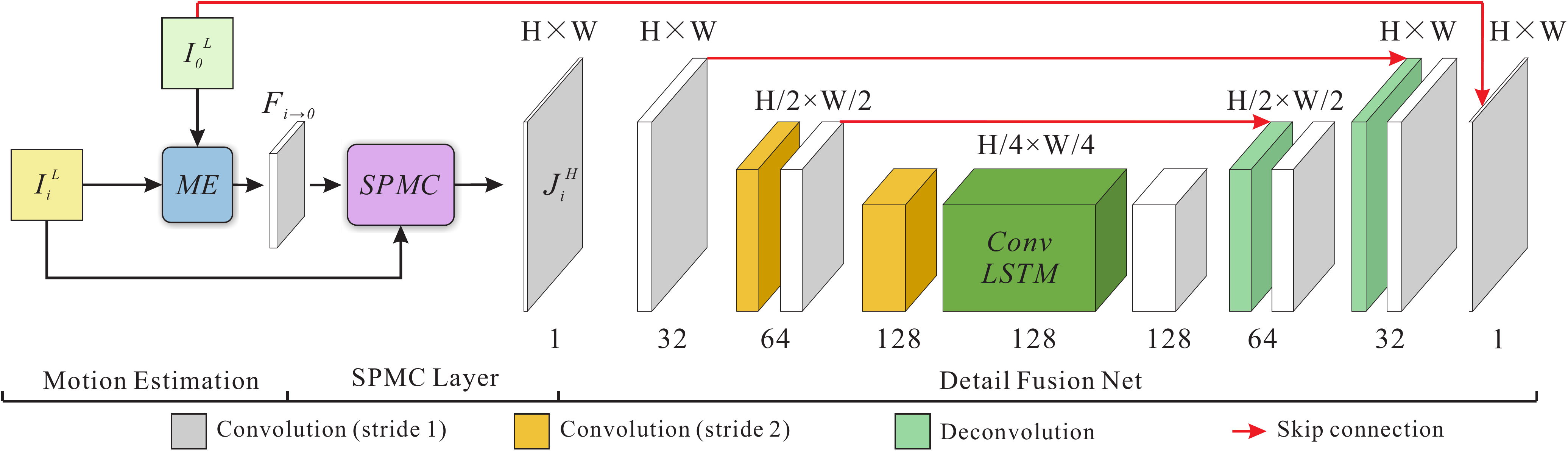}
\end{center}
\caption{\textbf{Our framework.} Network configuration for the $i$th time
step.}\label{fig:framework}
\end{figure*}

Our method takes a sequence of $N_F$ LR images as input and produces one HR image
$I^H_0$. It is an end-to-end fully trainable framework that comprises of three modules:
motion estimation, motion compensation and detail fusion. They are respectively
responsible for motion field estimation between frames; aligning frames by compensating
motion; and finally increasing image scale and adding image details. We elaborate on each
module in the following.

%-------------------------------------------------------------------------
\subsection{Motion Estimation}
The motion estimation module takes two LR frames as input and produces a LR motion field as
\begin{align}\label{eq:module_ME}
    F_{i\rightarrow j} = \mathbf{Net}_{ME}(I^L_i, I^L_j;\theta_{ME}),
\end{align}
where $F_{i\rightarrow j}=(u_{i\rightarrow j}, v_{i\rightarrow j})$ is the
motion field from frame $I^L_i$ to $I^L_j$. $\theta_{ME}$ is the set of module
parameters.

Using neural networks for motion estimation is not a new idea, and existing work
\cite{caballero2016real,fischer2015flownet,mayer2015large,yu2016back} already achieves
good results. We have tested FlowNet-S \cite{fischer2015flownet} and the motion
compensation transformer (MCT) module from VESPCN \cite{caballero2016real} for our task.
We choose MCT because it has less parameters and accordingly less computation cost. It
can process 500+ single-channel image pairs ($100\times 100$ in pixels) per second. The
result quality is also acceptable in our system.

\subsection{SPMC Layer}

\begin{figure}[t]
\centering
    \includegraphics[width=1.0\linewidth]{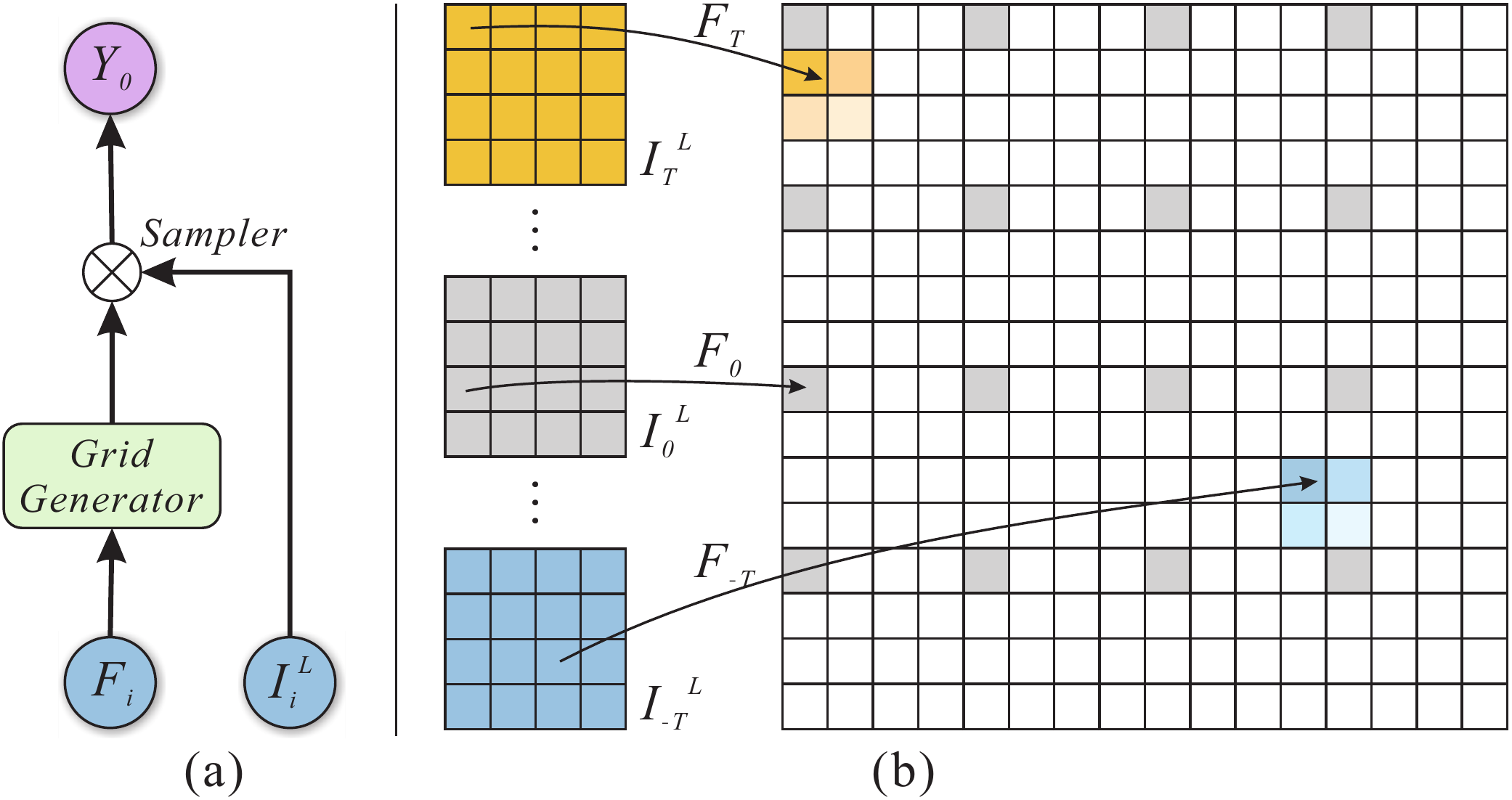}
    \caption{\textbf{Subpixel Motion Compensation layer} ($\times4$).
    (a) Layer diagram. (b) Illustration of the SPMC layer ($\times4$).
    }\label{fig:module_spmc}
\end{figure}

According to the analysis in Sec.~\ref{sec:analysis}, we propose a novel layer to utilize
sub-pixel information from motion and simultaneously achieve sub-pixel motion
compensation (SPMC) and resolution enhancement. It is defined as
\begin{align}\label{eq:spmc_layer}
  J^H = \mathbf{Layer}_{SPMC}(J^L,F;\alpha),
\end{align}
where $J^L$ and $J^H$ are input LR and output HR images, $F$ is optical flow used for
transposed warping and $\alpha$ is the scaling factor. The layer contains two submodules.

\vspace{-0.1in}
\paragraph{Sampling Grid Generator}
In this step, transformed coordinates are first calculated according to estimated flow
$F=(u,v)$ as
\begin{align}\label{eq:st_grid}
    \binom{x_p^s}{y_p^s} = W_{F;\alpha}\binom{x_p}{y_p} = \alpha\binom{x_p+u_p}{y_p+v_p},
\end{align}
where $p$ indexes pixels in LR image space. $x_p$ and $y_p$ are the two coordinates of
$p$. $u_p$ and $v_p$ are the flow vectors estimated from previous stage. We denote
transform of coordinates as operator $W_{F; \alpha}$, which depends on flow field $F$ and
scale factor $\alpha$. $x_p^s$ and $y_p^s$ are the transformed coordinates in an enlarged
image space, as shown in Fig.~\ref{fig:module_spmc}.

\vspace{-0.1in}
\paragraph{Differentiable Image Sampler}
Output image is constructed in the enlarged image space according to $x_p^s$ and $y_p^s$.
The resulting image $J_q^H$ is
\begin{align}\label{eq:st_sampler}
    J_q^H=\sum_{p=1} J_p^L M(x_p^s-x_q)M(y_p^s-y_q),
\end{align}
where $q$ indexes HR image pixels. $x_q$ and $y_q$ are the two coordinates for pixel $q$
in the HR grid. $M(\cdot)$ is the sampling kernel, which defines the image interpolation
methods (\eg bicubic, bilinear, and nearest-neighbor).

We further investigate differentiability of this layer. As indicated in
Eq.~(\ref{eq:spmc_layer}), the SPMC layer takes one LR image $J^L$ and one flow field
$F=(u,v)$ as input, without other trainable parameters. For each output pixel, partial
derivative with respect to each input pixel is
\begin{align}\label{eq:st_grad_J}
    \frac{\partial J^H_q}{\partial J_p^L} = \sum_{p=1} M(x_p^s-x_q)M(y_p^s-y_q).
\end{align}
It is similar to calculating partial derivatives with respect to flow field $(u_p,v_p)$
using the chain rule as
\begin{align}\label{eq:st_grad_uv}
\frac{\partial J^H_q}{\partial u_p}=\frac{\partial J^H_q}{\partial x_p^s}\cdot
\frac{\partial x_p^s}{\partial u_p} = \alpha\sum_{p=1} J_p^L M'(x_p^s-x_q)M(y_p^s-y_q),
\end{align}
where $M'(\cdot)$ is the gradient of sampling kernel $M(\cdot)$. Similar derivatives can
be derived for $\frac{\partial J_q}{\partial v_p}$. We choose $M(x)=\max(0,1-|x|)$, which
corresponds to the bilinear interpolation kernel, because of its simplicity and
convenience to calculate gradients. Our final layer is fully differentiable, allowing
back-propagating loss to flow fields smoothly. The advantages of having this type of
layers is threefold.
\begin{itemize}
  \vspace{-0.05in}
  \item This layer can simultaneously achieve motion compensation and resolution
    enhancement. Note in most previous work, they are
    separate steps (\eg backward warping + bicubic interpolation).
  \vspace{-0.1in}
  \item This layer is parameter free and fully differentiable, which can be
    effectively incorporated into neural networks with almost no additional cost.
  \vspace{-0.1in}
  \item The rationale behind this layer roots from accurate LR imaging model,
    which ensures good performance in theory. It also demonstrates good results in
    practice, as we will present later.
\end{itemize}

\subsection{Detail Fusion Net}
The SPMC layer produces a series of motion compensated frames $\{J^H_i\}$ expressed as
\begin{align}\label{eq:df_spmc}
  J^H_i = \mathbf{Layer}_{SPMC}(I^L_i,F_{i\rightarrow 0};\alpha).
\end{align}
Design of the following network is non-trivial due to the following considerations.
First, $\{J^H_i\}$ are already HR-size images that produce large feature maps, thus
computational cost becomes an important factor.

Second, due to the property of forward warping and zero-upsampling, $\{J^H_i\}$ is sparse
and majority of the pixels are zero-valued (\eg about 15/16 are zeros for scale factor
$4\times$). This requires the network to have large receptive fields to capture image
patterns in $J^H_i$. Using simple interpolation to fill these holes is not a good
solution because interpolated values would dominate during training.

Finally, special attention needs to be paid to the use of the reference frame. On the one
hand, we rely on the reference frame as the guidance for SR so that the output HR image
is consistent with the reference frame in terms of image structures. On the other hand,
over-emphasizing the reference frame could impose an adverse effect of neglecting
information in other frames. The extreme case is that the system behaves like a
single-image SR one.

\vspace{-0.1in}
\paragraph{Network Architecture} We design an encoder-decoder \cite{mao2016image} style
structure with skip-connections (see Fig.~\ref{fig:framework}) to tackle above issues.
This type of structure has been proven to be effective in many image regression tasks
\cite{liu2017voxelflow,mao2016image,su2016deep}. The encoder sub-network reduces the size
of input HR image to $1/4$ of it in our case, leading to reduced computation cost. It
also makes the feature maps less sparse so that information can be effectively aggregated
without the need of employing very deep networks. Skip-connections are used for all
stages to accelerate training.

A ConvLSTM module \cite{xingjian2015convolutional} is inserted in the middle stage as a
natural choice for sequential input. The network structure includes
\begin{align}\label{eq:df_convlstm}
  \mathbf{f}_i &= \mathbf{Net}_{E}(J^H_i;\theta_E)\nonumber\\
  \mathbf{g}_i,\mathbf{s}_i &= \mathbf{ConvLSTM}(\mathbf{f}_i, \mathbf{s}_{i-1};\theta_{LSTM})\\
  I^{(i)}_0 &= \mathbf{Net}_{D}(\mathbf{g}_i, \textbf{S}^E_i;\theta_D) + I^{L\uparrow}_0\nonumber
\end{align}
where $\mathbf{Net}_{E}$ and $\mathbf{Net}_{D}$ are encoder and decoder CNNs with
parameters $\theta_E$ and $\theta_D$. $\mathbf{f}_i$ is the output of encoder net.
$\mathbf{g}_i$ is the input of decoder net. $\mathbf{s}_{i}$ is the hidden state for LSTM
at the $i$th step. $\textbf{S}^E_i$ for all $i$ are intermediate feature maps of
$\mathbf{Net}_{E}$, used for skip-connection. $I^{L\uparrow}_0$ is the bicubic upsampled
$I^{L}_0$. $I^{(i)}_0$ is the $i$th time step output.

The first layer of $\mathbf{Net}_{E}$ and the last layer of $\mathbf{Net}_{D}$ have
kernel size $5\times5$. All other convolution layers use kernel size $3\times3$,
including those inside ConvLSTM. Deconvolution layers are with kernel size $4\times4$ and
stride 2. Rectified Linear Units (ReLU) are used for every conv/deconv layer as the
activation function. For skip-connection, we use SUM operator between connected layers.
Other parameters are labeled in Fig.~\ref{fig:framework}.

\subsection{Training Strategy}
Our framework consists of three major components, each has a unique functionality.
Training the whole system in an end-to-end fashion with random initialization would
result in zero flow in motion estimation, making the final results similar to those of
single-image SR. We therefore separate training into three phases.

\vspace{-0.1in}
\paragraph{Phase 1} We only consider $\mathbf{Net}_{ME}$ in the beginning of training. Since we
do not have ground truth flow, unsupervised warping loss is used as
\cite{liu2017voxelflow,yu2016back}
\begin{align}
\mathcal{L}_{ME} = \sum_{i=-T}^T \Vert I^L_i - \tilde{I}^L_{0\rightarrow i}\Vert_{1} +
\lambda_{1}\Vert \nabla F_{i\rightarrow 0}\Vert_{1}\label{eq:loss_flow},
\end{align}
where $\tilde{I}^L_{0\rightarrow i}$ is the backward warped $I^L_0$ according to
estimated flow $F_{i\rightarrow 0}$, using a differentiable layer similar to spatial
transformer \cite{jaderberg2015spatial}. Note that this image is in low resolution,
aligned with $I^L_i$. $\Vert \nabla F_{i\rightarrow 0}\Vert_{1}$ is the total variation
term on each $(u,v)$-component of flow $F_{i\rightarrow 0}$. $\lambda_1$ is the
regularization weight. We set $\lambda_1=0.01$ in all experiments.

\begin{figure*}[ht]
\begin{center}
\includegraphics[width=1.0\linewidth]{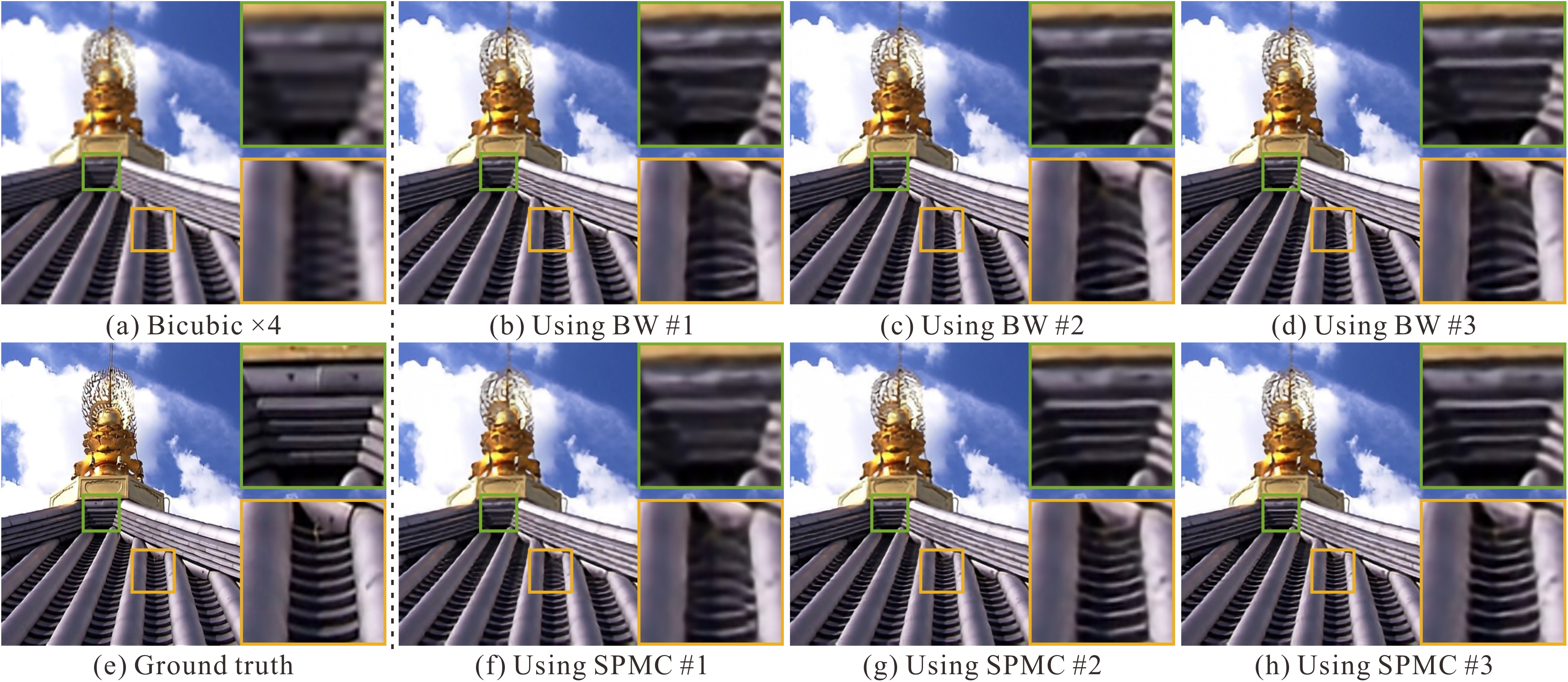}
\end{center}
\caption{\textbf{Effectiveness of SPMC Layer} (F3-$\times4$). (a) Bicubic $\times4$.
(b)-(d) Output for each time step using \textbf{BW}. (e) Ground truth. (f)-(h) Outputs
using \textbf{SPMC}.}\label{fig:exp_spmc}
\end{figure*}

\vspace{-0.1in}
\paragraph{Phase 2} We then fix the learned weights $\theta_{ME}$ and only
train $\mathbf{Net}_{DF}$. This time we use Euclidean loss between our estimated HR
reference frame and the ground truth as
\begin{align}
\mathcal{L}_{SR} = \sum_{i=-T}^T \kappa_i\Vert I^H_0 - I^{(i)}_0
\Vert^2_2\label{eq:loss_sr},
\end{align}
where $I^{(i)}_0$ is our network output in the $i$th time step, corresponding to
reference frame $I^L_0$. $\{\kappa_i\}$ are the weights for each time step. We
empirically set $\kappa_{-T}=0.5$ and $\kappa_{T}=1.0$, and linearly interpolate
intermediate values.

\vspace{-0.1in}
\paragraph{Phase 3} In the last stage, we jointly tune the whole system using the total loss as
\begin{align}
  \mathcal{L} = \mathcal{L}_{SR} + \lambda_{2}\mathcal{L}_{ME}\label{eq:loss_total},
\end{align}
where $\lambda_{2}$ is the weight balancing two losses.

%------------------------------------------------------------------------
\section{Experiments}
We conduct our experiments on a PC with an Intel Xeon E5 CPU and an NVIDIA Titan X GPU.
We implement our framework on the TensorFlow platform \cite{tensorflow2015-whitepaper},
which enables us to easily develop our special layers and experiment with different
network configurations.

\vspace{-0.1in}
\paragraph{Data Preparation}
For the super-resolution task, training data needs to be of high-quality without noise
while containing rich fine details. To our knowledge, there is no such publicly available
video dataset that is large enough to train our deep networks. We thus collect 975
sequences from high-quality 1080p HD video clips. Most of them are commercial videos shot
with high-end cameras and contain both natural-world and urban scenes that have rich
details. Each sequence contains 31 frames following the configuration of
\cite{liao2015video,liu2011bayesian,ma2015handling}. We downsample the original frames to
$540\times960$ pixels as HR ground truth using bicubic interpolation. LR input is
obtained by further downsampling HR frames to $270\times480$, $180\times320$ and
$135\times240$ sizes. We randomly choose 945 of them as training data, and the rest 30
sequences are for validation and testing.

\vspace{-0.1in}
\paragraph{Model Training}
For model training, we use Adam solver \cite{kingma2014adam} with learning rate of
$0.0001$, $\beta_1=0.9$ and $\beta_2=0.999$. We apply gradient clip only to weights of
ConvLSTM module (clipped by global norm 3) to stabilize the training process. At each
iteration, we randomly sample $N_F$ consecutive frames (\eg $N_F=3,5,7$) from one
sequence, and randomly crop a $100\times 100$ image region as training input. The
corresponding ground truth is accordingly cropped from the reference frame with size
$100\alpha\times 100\alpha$ where $\alpha$ is the scaling factor. Above parameters are
fixed for all experiments. Batch size varies according to different settings, which is
determined as the maximal value allowed by GPU memory.

We first train the motion estimation module using only loss $\mathcal{L}_{ME}$ in
Eq.~(\ref{eq:loss_flow}) with $\lambda_1=0.01$. After about 70,000 iterations, we fix the
parameters $\theta_{ME}$ and train the system using only loss $\mathcal{L}_{SR}$ in
Eq.~(\ref{eq:loss_sr}) for 20,000 iterations. Finally, all parameters are trained using
total loss $\mathcal{L}$ in Eq.~(\ref{eq:loss_total}), $\lambda_2$ is empirically chosen
as 0.01. All trainable variables are initialized using Xavier methods
\cite{glorot2010understanding}.

In the following analysis and experiments, we train several models under different
settings. For simplicity, we use $\times(\cdot)$ to denote scaling factors (\eg
$\times2$, $\times3$, and $\times4$). And F$(\cdot)$ is used as the number of input
frames (\eg F3, F5, and F7). Moreover, our ConvLSTM based DF net produces multiple
outputs (one for each time step), we use $\{\#1, \#2, \cdots\}$ to index output.

\subsection{Effectiveness of SPMC Layer}

We first evaluate the effectiveness of the proposed SPMC layer. For comparison, a
baseline model \textbf{BW} (F3-$\times4$) is used. It is achieved by fixing our system in
Fig.~\ref{fig:framework}, except replacing the SPMC layer with backward warping, followed
by bicubic interpolation, which is a standard alignment procedure. An example is shown in
Fig.~\ref{fig:exp_spmc}. In Fig.~\ref{fig:exp_spmc}(a), bicubic $\times4$ for reference
frame contains severe aliasing for the tile patterns. Baseline model \textbf{BW} produces
3 outputs corresponding to three time steps in Fig.~\ref{fig:exp_spmc}(b)-(d). Although
results are sharper when more frames are used, tile patterns are obviously wrong compared
to ground truth in Fig.~\ref{fig:exp_spmc}(e). This is due to loss of sub-pixel
information as analyzed in Section \ref{sec:analysis}. The results are similar to the
output of single image SR, where the reference frame dominates.

As shown in Fig.~\ref{fig:exp_spmc}(f), if we only use one input image in our method, the
recovered pattern is also similar to Fig.~\ref{fig:exp_spmc}(a)-(d). However, with more
input frames fed into the system, the restored images dramatically improve, as shown in
Fig.~\ref{fig:exp_spmc}(g)-(h), which are both sharper and closer to the ground truth.
Quantitative values on our validation set are listed in Table~\ref{tab:exp_ablation}.

\subsection{Detail Fusion vs. Synthesis}

\begin{figure*}[ht]
    \begin{center}
    \includegraphics[width=1.0\linewidth]{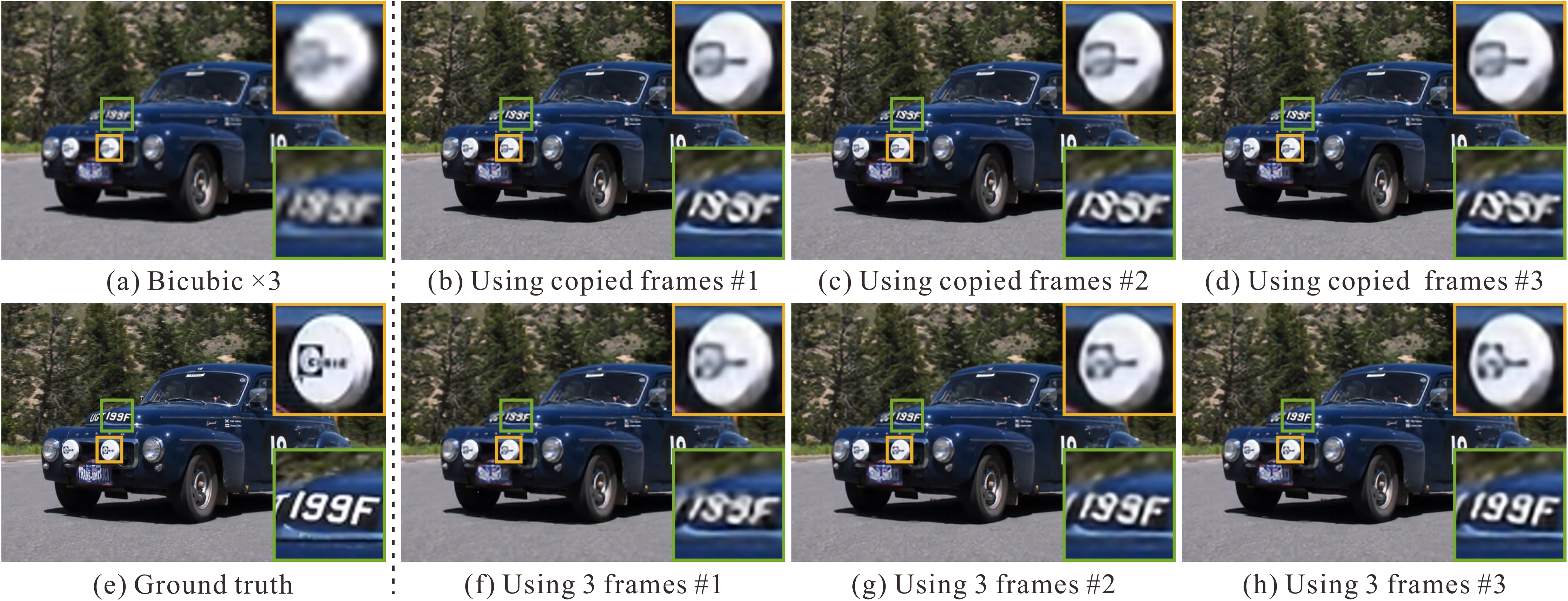}
    \end{center}
    \vspace{-0.1in}
    \caption{\textbf{SR using multiple frames} (F3-$\times3$). (a) Bicubic $\times3$.
            (b)-(d) Outputs for each time step using 3 reference frames that are the same. (e) Ground truth.
            (f)-(h) Outputs using 3 consecutive frames.}\label{fig:exp_truth_hall}
\end{figure*}

We further investigate if our recovered details truly exist in original frames. One
example is already shown in Fig.~\ref{fig:exp_spmc}. Here we conduct a more illustrative
experiment by replacing all input frames with the same reference frame. Specifically,
Fig.~\ref{fig:exp_truth_hall}(f)-(h) are outputs using 3 consecutive frames
(F3-$\times3$). The numbers and logo are recovered nicely. However, if we only use 3
copies of the same reference frame as input and test them on the same pre-trained model,
the results are almost the same as using only one frame. This manifests that our final
result shown in  Fig.~\ref{fig:exp_truth_hall}(h) is truly recovered from the 3 different
input frames based on their internal detail information, rather than synthesized from
external examples because if the latter holds, the synthesized details should also appear
even if we use only one reference frame.

\subsection{DF-Net with Various Inputs}

\begin{table}
    \center
    \caption{Performance of baseline models}\label{tab:exp_ablation}
    \vspace{0.1in}
    \scriptsize
    \begin{tabular}{c|c|c|c|c}
       \hline
       Model (F3)       & BW            & DF-Bic      & DF-0up       & Ours                  \\ \hline
       SPMCS ($\times4$)& 29.23 / 0.82  & 29.67 / 0.83& 29.65 / 0.83 & \textbf{29.69} / \textbf{0.84} \\ \hline
    \end{tabular}
    \vspace{-0.1in}
\end{table}

\begin{figure}[ht]
    \begin{center}
    \includegraphics[width=1.0\linewidth]{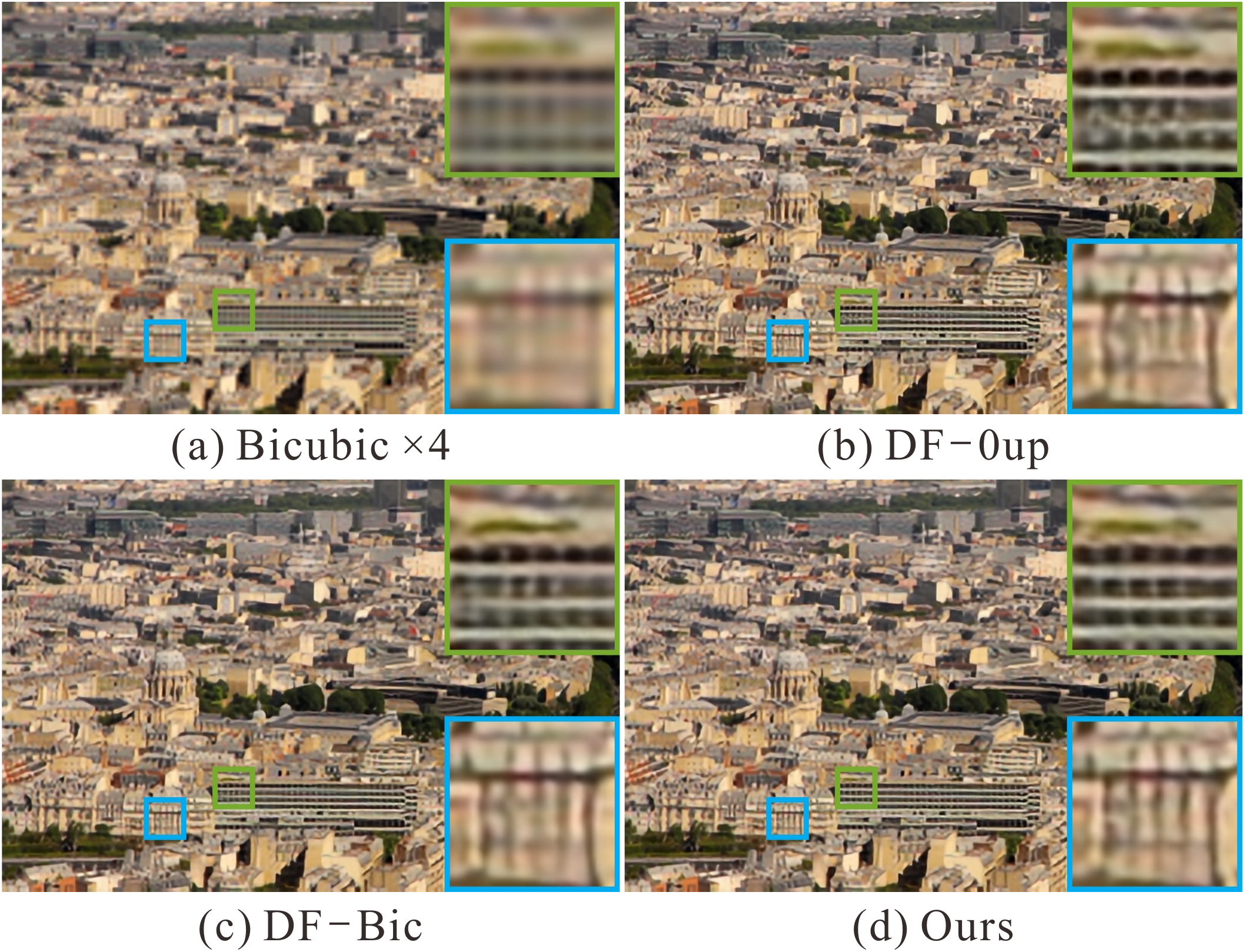}
    \end{center}
    \vspace{-0.1in}
    \caption{\textbf{Detail fusion net with various inputs.}}\label{fig:exp_dfnet}
    \vspace{-0.1in}
\end{figure}

Our proposed detail fusion (DF) net takes only $J^H_i$ as input. To further evaluate if
the reference frame is needed, we design two baseline models. Model \textbf{DF-bic} and
\textbf{DF-0up} respectively add bicubic and zero-upsampled $I^L_0$ as another channel of
input to DF net. Visual comparison in Fig.~\ref{fig:exp_dfnet} shows that although all
models can recover reasonable details, the emphasis on the reference frame may mislead
detail recovery and slightly degrade results quantitatively on the evaluation set (see
Table \ref{tab:exp_ablation}).

\subsection{Comparisons with Video SR Methods}

\begin{figure*}[ht]
\begin{center}
  \includegraphics[width=0.95\linewidth]{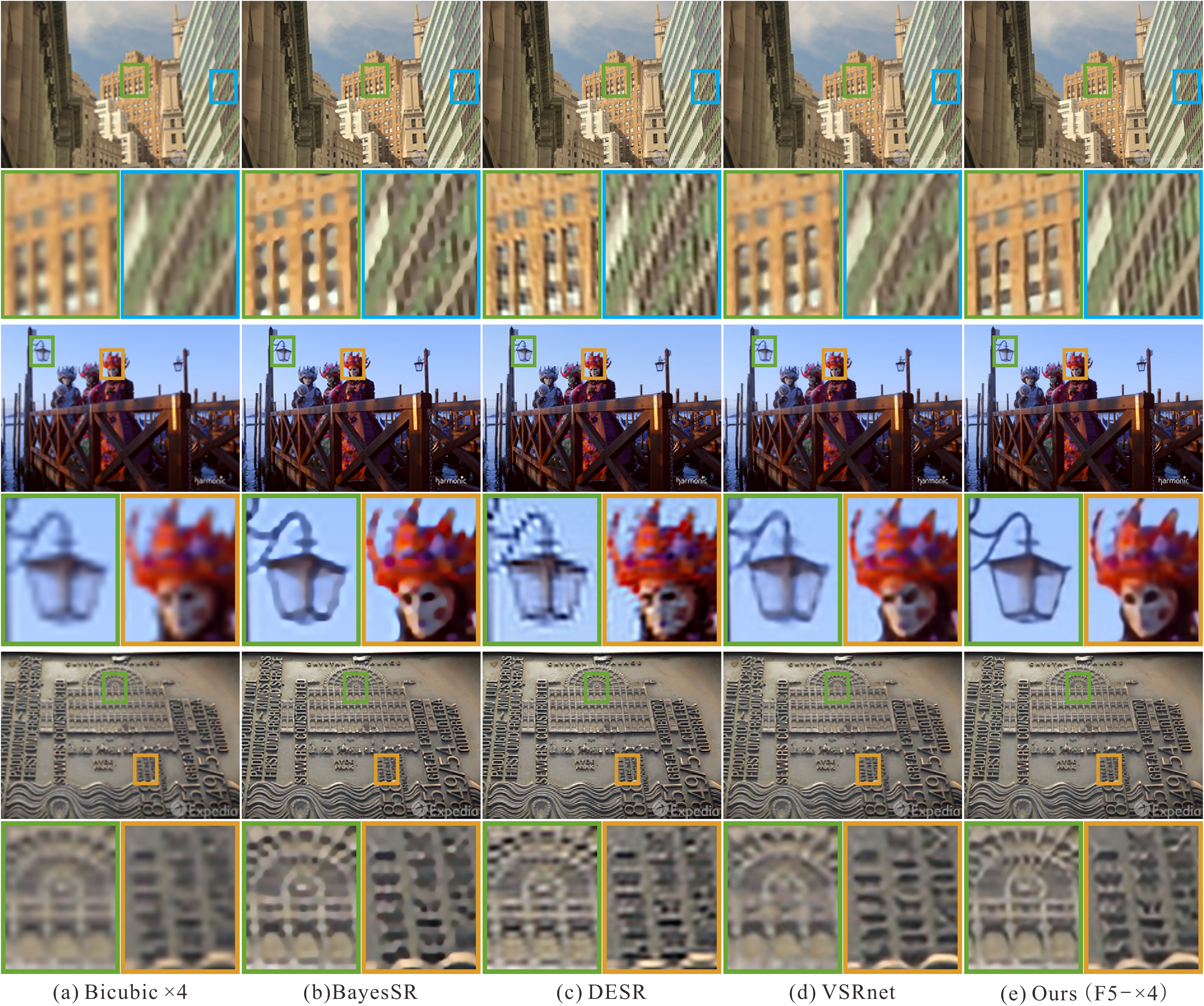}
\end{center}
   \vspace{-0.1in}
   \caption{Comparisons with video SR methods.}\label{fig:comp_videosr}
   \vspace{-0.2in}
\end{figure*}

We compare our method against previous video SR methods on the evaluation dataset. BayesSR
\cite{liu2011bayesian,ma2015handling} is viewed as the best-performing traditional method that
iteratively estimates motion flow, blur kernel, noise and the HR image. DESR
\cite{liao2015video} ensembles ``draft'' based on estimated flow, which makes it an
intermediate solution between traditional and CNN-based methods. We also include a recent
deep-learning-based method VSRnet \cite{kappeler2016video} in comparison. We use
author-provided implementation for all these methods. VESPCN \cite{caballero2016real}
did not provide code or pre-trained model, so we only list their reported PSNR/SSIM on
the 4-video dataset \textbf{VID4} \cite{liu2011bayesian}.
The quantitative results are listed in Table~\ref{tab:comparison_videoSR}. Visual comparisons are shown in Fig.~\ref{fig:comp_videosr}.

\begin{figure}[ht]
\begin{center}
%\fbox{\rule{0pt}{2in} \rule{1.0\linewidth}{0pt}}
   \includegraphics[width=1.0\linewidth]{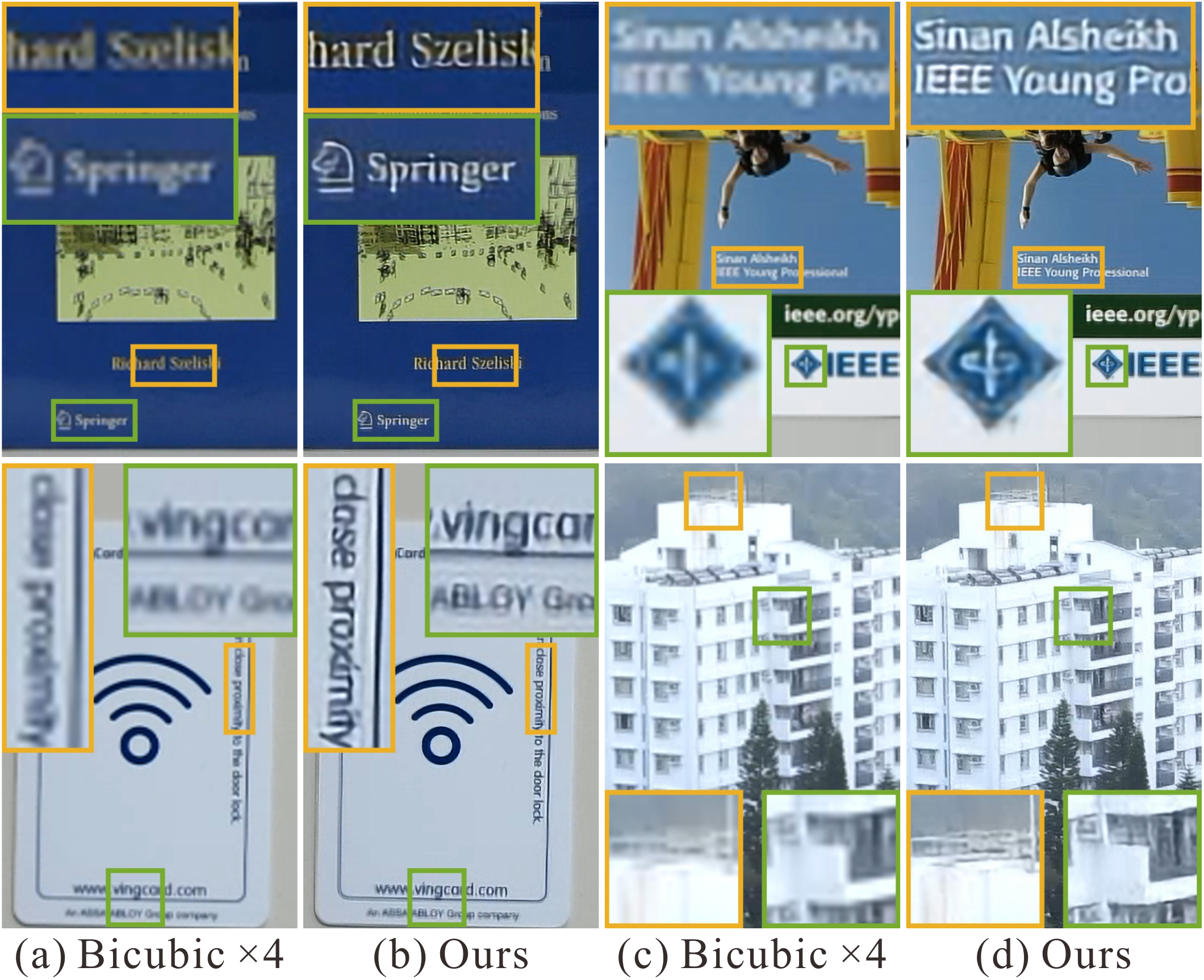}
\end{center}
   \vspace{-0.1in}
   \caption{\textbf{Real-world examples under configuration} (F7-$\times4$).}\label{fig:real_case}
   \vspace{-0.1in}
\end{figure}

\begin{table}
    \centering
    \caption{Comparison with video SR methods (PSNR/SSIM)}\label{tab:comparison_videoSR}
    \vspace{0.1in}
    \scriptsize
    \tabcolsep0.04in
    \begin{tabular}{c|c|c|c|c|c}
       \hline
       Method (F3)       & Bicubic     & BayesSR    & DESR         & VSRnet      & Ours (F3)\\ \hline
       SPMCS$\times2$    &32.48 / 0.92 &31.85 / 0.92& -            & 33.39 / 0.94 & \textbf{36.71} / \textbf{0.96} \\ \hline
       SPMCS$\times3$    &28.85 / 0.82 &29.42 / 0.87& -            & 28.55 / 0.85 & \textbf{31.92} / \textbf{0.90} \\ \hline
       SPMCS$\times4$    &27.02 / 0.75 &27.87 / 0.80& 26.64 / 0.76 & 24.76 / 0.77 & \textbf{29.69} / \textbf{0.84} \\ \hline\hline
       Method (F5)       & Bicubic     & BayesSR    & DESR         & VSRNet       & Ours (F5)\\ \hline
       SPMCS$\times2$    &32.48 / 0.92 &31.82 / 0.92& -            & 35.44 / 0.95 & \textbf{36.62} / \textbf{0.96} \\ \hline
       SPMCS$\times3$    &28.85 / 0.82 &29.55 / 0.87&-             & 30.73 / 0.88 & \textbf{32.10} / \textbf{0.90} \\ \hline
       SPMCS$\times4$    &27.02 / 0.75 &28.03 / 0.81& 26.97 / 0.77 & 28.35 / 0.79 & \textbf{29.89} / \textbf{0.84} \\ \hline\hline
       Method (F3)       & BayesSR     & DESR       & VSRNet       & VESPCN       & Ours (F3)\\ \hline
       Vid4$\times3$     &25.64 / 0.80 & -          & 25.31 / 0.76 & 27.25 / \textbf{0.84} & \textbf{27.49} / \textbf{0.84} \\ \hline
       Vid4$\times4$     &24.42 / 0.72 &23.50 / 0.67& 22.81 / 0.65 & 25.35 / \textbf{0.76} & \textbf{25.52} / \textbf{0.76} \\ \hline
    \end{tabular}
\end{table}

\subsection{Comparisons with Single Image SR}
\begin{figure}[ht]
\begin{center}
%\fbox{\rule{0pt}{2in} \rule{1.0\linewidth}{0pt}}
   \includegraphics[width=1.0\linewidth]{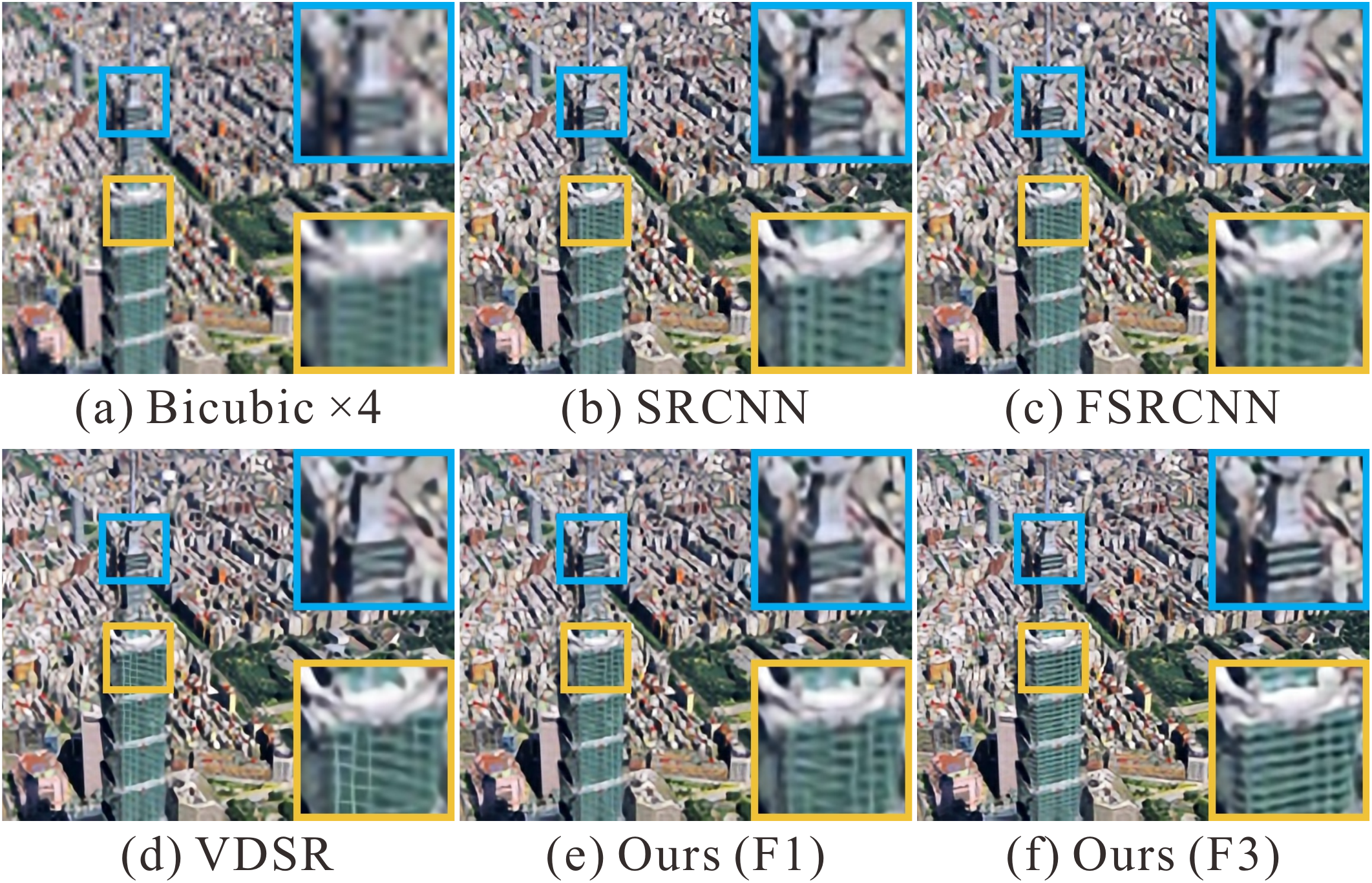}
\end{center}
    \vspace{-0.1in}
   \caption{\textbf{Comparisons with single image SR methods}.
            (a) Bicubic $\times4$. (b)-(d) Output from image SR methods. (e) Our result using 1 frame.
            (f) Our result using 3 frames.}\label{fig:comp_imagesr}
   \vspace{-0.1in}
\end{figure}

Since our framework is flexible, we can set $N_F=1$ to turn it into a single image SR solution.
We compare this approach with three recent image SR methods: SRCNN
\cite{dong2014learning}, FSRCNN \cite{dong2016accelerating} and VDSR \cite{kim2016VDSR},
on dataset \textbf{Set5} \cite{bevilacqua2012low} and \textbf{Set14}
\cite{zeyde2010single}. To further compare the performance of using multiple frames against
single, we compare all single image methods with our method under F3 setting on our
evaluation dataset \textbf{SPMCS}. The quantitative results are listed in Table
\ref{tab:comparison_singleSR}.

For the F1 setting on \textbf{Set5} and \textbf{Set14}, our method produces
comparable or slightly lower PSNR or SSIM results. Under the F3 setting, our method outperforms image SR methods by a large margin, indicating that
our multi-frame setting can effectively fuse information in multiple frames.
An example is shown in
Fig.~\ref{fig:comp_imagesr}, where single image SR cannot recover the tiled structure of
the building. In contrast, our F3 model can faithfully restore it.

\begin{table}
    \centering
    \caption{Comparison with image SR methods (PSNR/SSIM)}\label{tab:comparison_singleSR}
    \vspace{0.05in}
    \scriptsize
    \tabcolsep0.04in
    \begin{tabular}{c|c|c|c|c|c}
       \hline
       Method               & SRCNN        & FSRCNN         & VDSR           & Ours (F1)      & Ours (F3)  \\ \hline
       Set 5 ($\times2$)    & 36.66 / 0.95 & 37.00 / 0.96   & \textbf{37.53} / \textbf{0.96}  & 37.35 / \textbf{0.96}   & -   \\ \hline
       Set 5 ($\times3$)    & 32.75 / 0.91 & 33.16 / 0.92   & \textbf{33.66} / \textbf{0.92}  & 33.45 / \textbf{0.92}   & -   \\ \hline
       Set 5 ($\times4$)    & 30.49 / 0.86 & 30.71 / 0.88   & \textbf{31.35} / \textbf{0.88}  & 30.96 / 0.87   & -    \\ \hline
       Set 14 ($\times2$)   & 32.45 / 0.91 & 32.63 / 0.91   & \textbf{33.03} / \textbf{0.91}  & 32.70 / \textbf{0.91}   & -    \\ \hline
       Set 14 ($\times3$)   & 29.30 / 0.82 & 29.43 / 0.83   & \textbf{29.77} / \textbf{0.83}  & 29.36 / \textbf{0.83}   & -    \\ \hline
       Set 14 ($\times4$)   & 27.45 / 0.75 & 27.59 / 0.77   & \textbf{28.01} / \textbf{0.77}  & 27.57 / 0.76   & -    \\ \hline\hline
      SPMCS ($\times2$)     & 35.20 / 0.95 & 35.56 / 0.95   & 36.14 / 0.96   & 36.23 / 0.96   & \textbf{36.71} / \textbf{0.96}   \\ \hline
      SPMCS ($\times3$)     & 30.66 / 0.87 & 30.87 / 0.88   & 31.26 / 0.89   & 31.18 / 0.88   & \textbf{31.92} / \textbf{0.90}   \\ \hline
      SPMCS ($\times4$)     & 28.29 / 0.79 & 28.43 / 0.79   & 28.80 / 0.81   & 28.80 / 0.80   & \textbf{29.69} / \textbf{0.84}   \\ \hline
    \end{tabular}
    \vspace{-0.1in}
\end{table}

\subsection{Real-World Examples}
%\vspace{-0.1in}
The LR images in the above evaluation are produced though downsampling
(bicubic interpolation). Although this is a standard approach for evaluation
\cite{dong2014learning,dong2016accelerating,kappeler2016video,kim2016VDSR,liao2015video,liu2011bayesian},
the generated LR images may not fully resemble the real-world cases. To verify the
effectiveness of our method on real-world data, we captured four examples as
shown in Fig.~\ref{fig:real_case}. For each object, we capture a short video using a
hand-held cellphone camera, and extract 31 consecutive frames from it. We then crop a
$135\times 240$ region from the center frame, and use TLD
tracking~\cite{kalal2012tracking} to track and crop the same region from all other frames
as the input data to our system. Fig.~\ref{fig:real_case} shows the SR result of the
center frame for each sequence. Our method faithfully recovers the
textbook characters and fine image details using the F7-$\times4$ model. More examples
are included in our supplementary material.

\subsection{Model Complexity and Running Time}
Using our un-optimized TensorFlow code, the F7-$\times4$ model takes about $0.26s$ to
process 7 input images with size $180\times120$ for one HR output. In comparison,
reported timings for other methods (F31) are 2
hours for Liu \etal \cite{liu2011bayesian}, 10 min. for Ma \etal \cite{ma2015handling},
and 8 min. for DESR \cite{liao2015video}. VSRnet \cite{kappeler2016video} requires
$\approx$40s for F5 configuration. Our method can be further accelerated to 0.19s for F5 and 0.14s for F3.

%\if 0
%are compared with current
%state-of-the-arts.
%For fair comparison, we manually transfer neural network
%part of each methods \cite{dong2014learning,dong2016accelerating,kappeler2016video,kim2016VDSR,liao2015video}
%to the same TensorFlow platform (which are originally slow CPU-based/MATLAB codes) and runs on the same GPU.
%\xtao{maybe simply report their reported time is ok?}
%\begin{table}
%    \center
%    \caption{Running Time}\label{tab:running_time}
%    \vspace{0.1in}
%    \scriptsize
%    \begin{tabular}{c|c|c|c|c|c}
%       \hline
%       Method        & SRCNN   & VDSR  & FSRCNN & Liu \etal & Ours  \\ \hline
%       Time (s)      & 0   & 0 & 0   & 0   & 0    \\ \hline\hline
%       Method        & Ma \etal& FESR  & VSRNet & VESPCN    & Ours  \\ \hline
%       Time (s)      & 0   & 0 & 0   & 0   & 0  \\ \hline
%    \end{tabular}
%\end{table}
%\fi

%------------------------------------------------------------------------
\section{Concluding Remarks}
We have proposed a new deep-learning-based approach for video SR. Our method includes a
sub-pixel motion compensation layer that can better handle inter-frame motion for
this task. Our detail fusion (DF) network that can effectively fuse
image details from multiple images after SPMC alignment. We have conducted extensive
experiments to validate the effectiveness of each module. Results show that
our method can accomplish high-quality results both qualitatively and quantitatively, at the same time being flexible on scaling factors and numbers of input frames.

{\small
\bibliographystyle{ieee}
\bibliography{videosr}
}

\end{document}